# Right-wing German Hate Speech on Twitter: Analysis and Automatic Detection

Sylvia Jaki[1] and Tom De Smedt[2]

**Abstract**

Discussion about the social network Twitter often concerns its role in political discourse, involving the question of when an expression of opinion becomes offensive, immoral, and/or illegal, and how to deal with it. Given the growing amount of offensive communication on the internet, there is a demand for new technology that can automatically detect hate speech, to assist content moderation by humans. This comes with new challenges, such as defining exactly what is free speech and what is illegal in a specific country, and knowing exactly what the linguistic characteristics of hate speech are. To shed light on the German situation, we analyzed over 50,000 right-wing German hate tweets posted between August 2017 and April 2018, at the time of the 2017 German federal elections, using both quantitative and qualitative methods. In this paper, we discuss the results of the analysis and demonstrate how the insights can be employed for the development of automatic detection systems.

# 1 Introduction

During the 2015 European refugee crisis, nearly half a million refugees arrived in Germany (Eurostat 2017). This was more than double than the year before, with Germany's Chancellor Angela Merkel famously stating "Wir schaffen das" ('we can do this'). Most of the refugees were young men from Muslim countries, reportedly including some Islamic State militants disguised as asylum seekers (Reuters 2016). During this time, the country also witnessed a number of violent incidents, such as the 2015 New Year's Eve sexual assaults, where groups of young male refugees sexually assaulted women in Cologne, or the December 2016 Berlin attack, where an Islamic State militant drove a hijacked truck into a Christmas market, killing 12 and injuring 56. This sharply polarized the German sentiment towards refugees (YouGov 2016). In wake of these events, the 2017 German federal elections experienced a considerable rise in right-wing populism, with the political party Alternative für Deutschland (AfD) achieving a striking success with 12.6% of the votes.

[1] University of Hildesheim, Institut für Übersetzungswissenschaft und Fachkommunikation, `jakisy@uni-hildesheim.de`
[2] University of Antwerp, Computational Linguistics Research Group, `tom.desmedt@uantwerpen.be`

Social media such as Twitter and Facebook are believed to have played an important role in the electoral debate, mainly in propelling increasingly polarizing rhetoric. To illustrate this, Conover, Ratkiewicz, Francisco, Gonçalves, Menczer, & Flammini (2011) used a cluster analysis to examine 250,000 Twitter messages (tweets) posted in the lead-up to the US congressional midterm elections. They found a segregated partisan structure with few retweets between left and right-wing Twitter users. Davey & Ebner (2017:23) analyzed online right-wing partisan structure for the 2017 German federal elections, and report an increased connectivity between far-right activists and their followers. Such segregated 'echo chambers' of like-minded users lend themselves to more extreme sentiments than would be the case in face-to-face interactions (Colleoni, Rozza, & Arvidsson 2014). Kreißel, Ebner, Urban, & Guhl (2018:12) claim that only a small minority of German Facebook users (5%) are responsible for most of the online hate speech. These users can mostly be traced to AfD and the Austrian Identitäre Bewegung (IB).

The German Network Enforcement Act (*Netzwerkdurchsetzungsgesetz*, NetzDG) now forces social media networks to systematically review reported hate speech (see section 3.1). For the most part, the response by IT companies has been cautious. Twitter has argued that 'no magic algorithm' exists for detecting extremist content (Twitter 2016), and Facebook has stated that it will take 5 to 10 years to develop reliable AI detection systems (Reuters 2018). Apart from the technical challenges with automatic hate speech detection, regulators also have to decide to what extent they leave the challenge up to AI, especially since detection algorithms have also been criticized for promoting censorship and/or perpetuating bias.

In this paper, we present our analysis of online hate speech following the 2017 German federal elections, and an attempt for the automatic detection of such discourse on Twitter, a popular microblogging platform with short, publicly visible messages (280 characters). We first discuss the steps taken to collect relevant data (section 2) before we provide an overview of what constitutes right-wing German hate speech from a legal and linguistic perspective (sections 3 and 4). We then outline the technical details of our machine learning approach (section 5), and discuss our findings with respect to ethics and future applications (section 6).

# 2 Methods & Materials

## 2.1 Data Collection

Between August 2017 and April 2018, we collected over 50,000 hateful tweets from 100+ right-wing German Twitter users, using the Pattern toolkit (De Smedt & Daelemans 2012) and the Twitter API. To collect the data, we manually identified 112 subversive Twitter profiles and then automatically collected their tweets. Subversive profiles are those that (1) post tweets containing racial slurs, profanity, and violent rhetoric, (2) do so repeatedly and consistently, and (3) indicate far-right ideology, either directly (i.e., in the profile description) or indirectly (e.g., through coded language or visual metaphors). For each profile, we started by searching Twitter for derogatory keywords, such as *Neger* ('nigger'), and examined the search results. If a profile posted vigorously about refugees, and had three or more tweets that could be considered offensive or illegal (see section 3.1 for a definition of these terms), it was retained. About 50% of these profiles posted 250 tweets or less in the 9-month period. About 25% posted 750 or more. One profile posted nearly 3,000 tweets (~10/day) and was banned from Twitter. Many of these profiles (about 65%) mention each other extensively, which supports Conover et al (2011) and Davey & Ebner (2017).

We also collected 50,000 'safe' tweets for comparison. These include about 20,000 tweets by more than 35 elected German politicians (evenly from all elected parties), who we expected to talk about relevant topics without posting offensive content. They also include 30,000 German tweets by as many Twitter users, talking about family, work, holidays, and so on. By comparing both datasets side-by-side using statistical methods such as feature selection, we can reveal stylistic cues that uniquely identify hate speech.

The EU asserts that IT companies now remove over 70% of reported hate speech (EU Memo 18-262, January 2018),[3] although there is no accepted procedure to decide what kind of language is hateful (cf. also section 3). In our case, 90% of the profiles used in our analysis were still online in April 2018. In the 9-month period from August to April, we observed less than 10 suspensions out of 100+ subversive profiles.

---

[3] http://europa.eu/rapid/press-release_MEMO-18-262_en.htm

In compliance with the EU General Data Protection Regulation (GDPR),[4] no personal data was retained after the analysis except for the tweet IDs, which can be used to reconstruct the dataset with the Twitter API if necessary.

## 2.2 User Profiles

Over 25% of the profiles in the dataset contain neo-Nazi or extremist right-wing cues in their username, for instance, neo-Nazi ciphers and abbreviations. Five usernames end in -88, a substitution cipher for HH or *Heil Hitler* (Turner-Graham 2015). Two end in -18 (AH, *Adolf Hitler*) and another two in -59 (EI, *Eil Itler*). One username contains two lightning emoticons for SS, which in combination with the historic pictures posted by the user is an obvious reference to Third Reich SS. Two usernames contain an IB prefix, referring to the Austrian Identitäre Bewegung (Hentges, Kökgiran, & Nottbohm 2015). Three usernames contain references to Nazi occultism, e.g., Thule Society or the Holy Grail (Goodrick-Clarke 1993). Four usernames can be associated with Teutonic mythology, mentioning Norse gods such as Thor and Freya, or Norse mythological creatures such as frost giants (*Hrimthursen*). Three usernames mention the Black Front, the Aryan race, or nerve gas, and two profiles support neo-Nazi music bands and hooliganism. Other cues of right-wing extremism are racial slurs, used in three usernames, and grandiose proclamations such as being a 'Guardian of Germany' or a 'Defender of the West' (another three).

Some of these cues could be coincidental, especially the ciphers. But this is ruled out by examining the user's profile picture and profile content. Several users also have a profile picture with a *völkisch* rune (Mees 2008), crusader heraldry (Takács 2015), Hitler parodies, portraits of or quotes by Nazi frontmen, or various other explicit imagery, such as German landmarks burning in flames, rioting, skulls, wolf packs, or crusader knights.

Most profile descriptions contain anti-refugee statements (e.g., *Gegen Islamismus, Islamkritiker*), pro-National Socialism (*National Sozialistisch, NS Jetzt*), or pro-white power statements (*White Power Worldwide*). Profile descriptions are particularly revealing about the image users want to project about themselves. Some state their mission explicitly, either by proclaiming to be a member of (for example) the far-right network Reconquista Germanica, or with statements like "Stoppt die Islamisierung Deutschlands" ('stop the islamization of Germany'). Some users give less explicit hints about how to interpret their tweets: "Politisch inkorrekt" ('politically incorrect') or "Polizei- und Nachrichtenmeldungen über übergriffige Flüchtlinge und Migranten" ('information about criminal refugees and immigrants from police and media reports'). Others highlight ideological preferences ("ungläubig, islamophob, populistisch" 'irreligious, islamophobic, populist') or demonstrate education ("Akademiker ohne ausgeprägtes Gutmenschentum" 'academic without pronounced do-goodiness'). Some users add famous quotes that can be interpreted as ideological statements, such as "Wenn Unrecht zu Recht wird, wird Widerstand zur Pflicht!" ('when injustice becomes law, resistance becomes duty'; Bertolt Brecht).

---

[4] https://ec.europa.eu/info/law/law-topic/data-protection_en

Despite some users' self-proclaimed education, language errors are common in the dataset. Many users post actively, so misspellings can be attributed to typos. Many users also do not use capitalization, but this is a known feature of the medium: it is common on social media to write without punctuation, especially when upset. But there are also numerous errors that indicate a foreign speaker, which becomes obvious in tweets with major syntactic errors,[5] or with specific types of grammatical errors such as gender.[6]

This finding is interesting insofar as it indicates that German hate speech on Twitter is not necessarily restricted to native German speakers. However, errors are certainly not restricted to non-native speakers, but could also indicate a low language proficiency. This is apparent in minor grammatical errors explicable by the transfer of grammatical structures from spoken non-standard varieties to a written context, as in "Wem wundert das?" (dative case of *wer* instead of accusative case).

---

[5] In "Unter Burka jede möglichen Figur kann sich verstecken. Daher Burkaträgerinnen sind undefinierbare Objekte *g* wie UFO 's", *Burka* is missing an article and the verbs *kann* and *sind* should come directly after the adverbial and the adverb.

[6] In "ICH wünsche Trump langes Leben und genau die Gegenteil wünsche ich zum Merkel und die politische Bande; Muss reserviert für Die ganze Mafia Bande !! Der Zeit wird kommen!!!! Brauchen nur neues Wahl!", an article is missing before *langes*, there is an error with the passive in *reserviert*, the pronoun *wir* is dropped before *brauchen*, and gender errors (*das Gegenteil*, *die Zeit*, etc.).

# 3 Defining Hate Speech

There is no universal definition of hate speech, perhaps as argued by Berger (2018) because of the far-right's proximity to political power. In general terms, hate speech can be seen as the communicative production of human inferiority (Sponholz 2018:48). It is a heterogeneous phenomenon that does not necessarily always involve attitudes of hatred (Meibauer 2013:3; Brown 2017:432). In legal terms, the definition of hate speech depends on the regulations in each individual country. This is a challenge in light of removing online content. For example, the Encyclopedia of the American Constitution defines hate speech as "any communication that disparages a person or a group on the basis of some characteristic such as race, color, ethnicity, gender, sexual orientation, nationality, religion, or other characteristic" (Nockleby 2000). It is rarely penalized in the US, where freedom of speech is absolute and protected by the First Amendment (Cohen 2014:247). Exceptions can include defamation, fighting words ("personal epithets hurled face-to-to face at another individual which are likely to cause the average addressee to fight", Stone 1994:80), threats of unlawful harm, and incitement to violence or other unlawful actions (cf. Fisch 2002).

In Germany, hate speech is more likely to be penalized. Its legislation is representative of the EU approach to hate speech regulation, which counterbalances freedom of expression with values such as human dignity (Cohen 2014:238). More concretely, in case of a conflict, US legislation tends to give precedence to freedom of speech while EU nations tend to give precedence to human dignity. Additionally, EU legislation also penalizes hate speech that targets groups, whereas US legislation is more restricted to penalizing speech that targets individuals (Cohen 2014:238f). This is also reflected in the EU's Code of Conduct on countering illegal hate speech online, which defines hate speech as "the public incitement to violence or hatred directed to groups or individuals on the basis of certain characteristics, including race, colour, religion, descent and national or ethnic origin".[7]

German legislation in particular shows a "heightened sensitivity toward individual dignity and the potential harms of violating these values", which evolved from past incidences during the Nazi regime (Cohen 2014:240). This includes, specifically, the prohibition of Holocaust denial (ARTICLE 19 2018:21). In general, §5(1) of the German *Grundgesetz* protects freedom of speech and prohibits censorship, but it is overruled when personal dignity is in danger (§5(2)). Infringements of personality rights are illegal and specified in German criminal law (*Strafgesetzbuch*, StGB), such as incitement to criminal behavior (*Öffentliche Auf- forderung zu Straftaten*, §111 StGB), incitement of the masses (*Volksverhetzung*, §130 StGB), insult (*Beleidigung*, §185 StGB), defamation (*Üble Nachrede*, §186 StGB), slander (*Verleumdung*, §187 StGB), coercion (*Nötigung*, §240 StGB), and intimidation (*Bedrohung*, §241 StGB) (Puneßen 2016). Also relevant is defamation or slander of a politician (*Üble Nachrede und Verleumdung gegen Personen des politischen Lebens*, §188 StGB) (Griffen 2017).

---

[7] http://europa.eu/rapid/press-release_MEMO-18-262_en.htm

Tweets qualifying as an offence as per StGB can be legally pursued. It is this kind of illegal hate speech (as well as fake news) that the Network Enforcement Act (NetzDG), effective in Germany as of October 1, 2017, pertains to. Reported illegal content (*Beschwerden über rechtswidrige Inhalte*, §3 NetzDG) must now be systematically reviewed within 24 hours. One type of content that is problematic here is satire (irony, ridicule), for which various cases of overblocking have been observed (ARTICLE 19 2018:24).

A common argument against NetzDG is that it constitutes censorship, by restricting freedom of speech. However, NetzDG only fosters the removal of content that was not protected by freedom of speech even before its introduction. In the dataset, the number of clearly illegal messages is relatively small. Table 1 shows a sample of tweets that qualify for discussion in terms of the above-mentioned articles.

| CASE | EXAMPLE |
|---|---|
| INCITEMENT §111 StGB | DE: *Findet diese Drecksau und entmannt sie an Ort und Stelle.* |
| | EN: Find the pig and kill him. |
| INCITEMENT §130 StGB | DE: *Wir müssen uns wehren sonst sind wir bald Fremde.* |
| | EN: We have to fight back now or we will soon be foreigners in our own country. |
| INSULT §185 StGB | DE: *Verpiss dich aus Deutschland du scheiss Kreatur.* |
| | EN: Piss off and leave Germany, you nasty creature. |
| SLANDER §186 StGB | DE: *Steigt Motumbo aus der Bahn, ist er sicher schwarz gefahn.* |
| | EN: Mutombo will surely have no train ticket. |
| SLANDER §188 StGB | DE: *Schulz muss erst mal das Saufen aufgeben, zum Wohle seiner Partei.* |
| | EN: Schulz needs to give up drinking for the sake of his party. |

**Table 1**. Examples of German hate speech compared to German criminal law.

Many instances of online group-focused enmity are currently in limbo between legally acceptable and illegal (Bundeszentrale für politische Bildung 2017). For example, while StGB clearly differentiates between simply spreading defamatory information, and knowing that this information is incorrect (slander), the distinction is harder to make in reality. With respect to European hate speech legislation, "the application and interpretation of the existing criminal provisions on 'hate speech' is generally inconsistent" (ARTICLE 19 2018:6).

We would argue to understand hate speech beyond its purely legal application, not only because of the lack of a clear-cut distinction between legal and illegal, but also because content moderation in social networks extends beyond what is illegal. For example, Facebook is known to apply its own rules, which only in part correspond to what should be removed by NetzDG.[8] Hate speech is now widely studied and considered a problem not so much because of the (relatively few) illegal cases, but because it is an indicator of society's polarization. The analysis thus includes "'[h]ate speech' that is lawful [...], but which nevertheless raises concerns in terms of intolerance and discrimination" (ARTICLE 19 2018:9) – tweets that are not illegal but 'merely' offensive, i.e., constituting either an insult or abuse. According to Ruppenhofer, Siegel, & Wiegand (2018:1f), this means messages that convey disrespect and contempt, by ascribing qualities to people that portray them as unworthy or unvalued (insult), or constitute group-based offense by ascribing a social identity to a person that is judged negatively by society or part of society (abuse).

Many tweets in the dataset involve a tendentiously framed dissemination of news, as people interpret news by connecting events to personal experience and world views (Maireder & Ausserhofer 2014:307). By framing we mean 'select[ing] some aspects of perceived reality and mak[ing] them more salient in a communicating text' (Entman 1993:52). For example, a crime committed by a single refugee may be framed as something that all refugees do all the time. This interpretation is often explicitly stated, as in "Dieser Mehrfach-Straftäter ist der Beweis: Multikulti funktioniert... Nicht!" ('this repeat offender proves that multiculturalism does not work'). Here, one criminal act is used to imply that multiculturalism automatically leads to an increased crime rate. The reported event may either be real news or fake news.

---

[8] https://www.facebook.com/communitystandards

**Real news.** Many tweets are factual, as in "Dieser Mann schnitt seiner zweijährigen Tochter die Kehle durch" ('man cuts throat of baby girl'), referring to a news article about a Pakistani man. Citing crime reports is not hate speech, but many users in the dataset do so repeatedly and exclusively about people of foreign origin. On average, a profile in the dataset posts a tweet like this every three hours. By comparison, German politicians posted a tweet every six hours during their electoral campaign. The users in the dataset provide a continuous stream of negative tweets implicating foreigners in criminal and violent activities.

**Fake news.** This stream of crime reports often includes hearsay and prejudice. In the April 2018 Münster attack, a man drove a van into the crowd in front of a restaurant, killing 4 and injuring more than 20. One AfD politician responded by tweeting "Wir schaffen das!", suggesting that the incident involved refugees, but the perpetrator turned out to be a German man with psychiatric problems. Many users in the dataset tweet about non-existent jihadist attacks, in this case for example: "Muslimattacken mit Autos lohnen wieder. Deutsche wacht endlich auf!" ('Muslim car attacks pay off again. Wake up, Germans!'). The wake-up call is reminiscent of the Third Reich slogan *Deutschland Erwache*.

Such messages constitute a new form of propaganda. Generally, propaganda can be defined as a "deliberate, systematic attempt to shape perceptions, manipulate cognitions, and direct behavior to achieve a response that furthers the desired intent of the propagandist" (Jowett & O'Donnell 2019:6). In today's digital society, it is increasingly associated with grassroots activities and fake news (Jowett & O'Donnell 2019:3-5). Most examples we found are not systematic in the sense that there is some elite conspiracy, but, as Davey & Ebner (2017) show, online activities of right-wing extremist networks are now well-organized, raising security concerns (Europol 2019). Farkas & Neumayer (2018:3f) point to an early distinction by Ellul (1965) between vertical and horizontal propaganda. Although propaganda in a digital context cannot be fully grasped by this binary distinction (Farkas & Neumayer 2018:7), the notion of horizontal propaganda, i.e., emerging from small groups who cooperate on the basis of a shared ideology, seems relevant for right-wing extremism on social media.

In conclusion, our working definition of hate speech includes illegal utterances but also legal cases that are nonetheless offensive, or those used for propaganda. As argued by Waseem, Davidson, Warmsley, & Weber (2017), abuse can be described in two broad dimensions: the specificity of the target and the degree of explicitness. They discern four categories useful for our work: (1) directed and explicit, i.e., directed towards a specific target and unambiguous, such as slurs, (2) directed and implicit, i.e., targeted but ambiguous, such as sarcasm, (3) generalized and explicit, i.e., not targeting anybody specific, and (4) generalized and implicit. Most of the dataset is generalized hate speech, which is typical of political communication (directed hate speech being typical of cyberbullying).

## 4 Analysis

To establish what online hate speech looks like on the linguistic level, and how it is used in context, we used a combination of qualitative approaches with quantitative techniques from Natural Language Processing (NLP), since the two should go hand-in-hand in research about political communication on Twitter (Pal & Gonawela 2017). For the qualitative part, we first examined a random subsample of 2,000 tweets. We then compared our insights to supporting quantitative evidence. We focused on a selection of self-chosen aspects of hate speech: the targets of hateful tweets and the designations used to describe them (4.1), lexis in general (4.2), linguistic creativity (4.3), and speech acts (4.4).

At first glance, not all tweets in the qualitative sample are hateful. There are also non-hateful messages for example extending good wishes, as in "Schönen Urlaub dir/euch!" ('have a nice holiday'). Non-political posts are less likely to be offensive, with the exception of football tweets, which perhaps is indicative of the connection between right-wing extremism and football ultras (Pilz n.d.). But non-political tweets are rare in the sample (<10%), suggesting that Twitter is used more purposefully for propaganda (section 3).

The majority of tweets relate to political parties and political ideology, or target immigration and purported crimes by refugees. Note that the examples presented are parts of tweets or tweets as a whole, including grammar or spelling mistakes. Some tweets have been redacted to protect the users' identity, while retaining the gist of their message.

## 4.1 Targeted Groups

While hate speech is disparaging by nature, disparaging tweets do not always necessarily use disparaging language, e.g., *der friedliche Islam* ('peaceful Islam') is used ironically. However, they do often resort directly to disparaging words, such as insults or negatively connoted words. These linguistic cues are very useful to expose who is targeted by politically motivated hate tweets, namely immigrants (~26%), political opponents (22%), and other German voters or Germany as a whole (13%). The dataset includes other targets besides these three, for example women (especially feminists) and homosexuals, but less frequently.

**Immigrants** and refugees in particular are considered a danger to Germany's safety, which is why hate speech targeting this group is abundant (cf. Geyer 2017). Immigrants are designated as *Nafris*,[9] *Invasoren* ('invaders'), *Asyltouristen* ('asylum tourists'), *Merkel-Gäste* ('Merkel's guests'), *Mob*, and occasionally also as *Illegale* ('illegal residents'), *Wohlstandsflüchtlinge* ('fortune seekers'), *Bunte* ('multi-colored people'), and as *Zudringlinge* ('intruders'). They are portrayed as being *kriminell* ('criminal'), *unterentwickelt* ('primitive'), *Müll* ('garbage'), *Abschaum* ('scum'), *Pack* ('vermin'), *Parasiten* ('parasites'), and *Gesindel* ('rabble'). Expressions such as *Musel*, *Salafistenschwester*, *Kampfmuslimas*, *Burka-Frauen*, or *Vollbärte* ('full beard') are used in relation to Muslims, *Mutombo*, *Bongo*, or *Kloneger* ('toilet nigger') in relation to African immigrants.

**Political opponents** include individuals from the entire political spectrum, but most notably SPD politicians like Martin Schulz (*Arschkriecher* 'brown-noser'), Heiko Maas (*Vollpfosten* 'idiot'), and Ralf Stegner (*Einzeller* 'single-cell organism'), or CDU politicians Stanislaw Tillich (*Bodensatz* 'dregs') and, in 8% of tweets, Angela Merkel (*Volksverräterin* 'betrayer of the people', *blöde Kuh* 'stupid cow'). Generally, opponents include all left-wing politicians and parties (*Verbrecher* 'criminals', *Sozi Clowns* 'socialist clowns', *SPD Heuchler* 'Social Democrat hypocrites', *linkes Faschistenpack* 'left-wing fascist vermin', *Grünfaschisten* 'green fascists'), which are portrayed as being *dumm* ('dumb'), *gehirnamputiert* ('brainless'), and *neokolonial* ('neocolonial'). They spread *Gelaber* ('nonsense') and *Lügenpropaganda* ('propagandist lies') in concert with the *Lügenpresse* ('fake news media').[10]

---

[9] *Nafri* is short for *Nordafrikaner* ('North African') or *Nordafrikanischer Intensivtäter* ('North African repeat offender'), and used in police jargon before it spread to general language.

[10] *Lügenpresse* was elected Non-word of the year 2014 in Germany, and has a long history (cf. Amendt 2015; Heine 2015). It dates back to mid-19th century but its usage increased greatly after 1914. It was used in Nazi propaganda by Joseph Goebbels.

**Voters** that have a positive attitude towards refugees, and mainstream voters in general, are designated as *Gutmenschen* ('do-gooders') or *Traumtänzer* ('dreamers').[11] By consequence, Germany is perceived as being in decline and referred to as *Buntland* ('rainbow nation'), *Dummstaat* ('idiot nation'), *PlemPlemLand* ('nation of fools'), or *Schandland* ('nation of shame'), and Berlin as a *Bundeskloake* ('cesspit') and *Dreckloch* ('shithole'). Its citizens are perceived to be *Idioten* ('idiots') and *verblödet* ('stupid').

## 4.2 Lexis

### 4.2.1 Word Bias

We want to know what vocabulary constitutes hate speech. In statistics, the chi-square test estimates how likely it is that an observed distribution is due to chance. For example, is it due to chance that the hate speech dataset contains a lot of derogatory words, or do all forms of online communication use derogatory words? In NLP, the chi-square test is used as a method for feature selection (Liu & Motoda 2007), to automatically expose relevant keywords.

We can count the number of times that a word occurs in all tweets (50,000 hate + 50,000 safe), and then observe if it occurs more often in the hate subset. We would expect function words such as *der*, *die*, *das* to occur equally often in any kind of text, since they are usually grammatically required. Contrarily, we would expect content words such as *Muslimenhorden* to be absent from the usual tweets that people post about their pets or cooking skills. As it turns out in this case, thousands of words (~4,500) are significantly biased ($p<0.05$) and occur much more often in the hate speech data, including racial slurs, ideological insults, and verbs expressing aggression.[12]

Table 2 shows a sample of biased words, with the total number of times that they appear in both datasets (#), sorted by the likelihood that they appear in the hate data (%) as opposed to the safe data, along with an example. We can also use such words as cues to automatically predict whether a text that we have not seen before is hateful (see section 5).

---

[11] *Gutmensch* was elected Non-word of the year 2015 in Germany, and dates back to 1922, when it was coined by columnist Karl Heinz Bohrer (Zifonun 2016:26) to describe people "that do not say everything that they want to say" to avoid provocation.

[12] Top 1,000 biased words: https://docs.google.com/spreadsheets/d/1JCYspKqgNx0PSqy5YZ03t63ks1eoE1p5KzZ9O5BUkQU

| WORD | # | % HATE | EXAMPLE |
|---|---|---|---|
| Nafri | 21 | 99% | *nafri müll raus aus deutschland!!!* |
| Salafisten | 42 | 98% | *Soso, die Salafistenschwester ruft also zum Kampf auf* 😏 |
| sexuell | 57 | 96% | *islamische Prediger mit gewisser Neigung zu sexueller Gewalt* |
| Horden | 15 | 93% | *bewaffnet euch vor den heranstürmenden Horden* |
| Volk | 518 | 90% | *fahrt zur Hölle, das Deutsche Volk lässt sich nicht mehr verarschen* |
| Gutmenschen | 160 | 90% | *Gutmenschen ist es scheissegal wieviele Mädchen sterben* |
| brutal | 72 | 89% | *brutale Rumänen überfallen Senioren* |
| Patrioten | 60 | 88% | *Das ist eine Kampfansage gegen uns Patrioten. Es wird ernst.* |
| töten | 103 | 87% | *ich würde Töten–bei Gott das schwöre ich!* |
| Kopftuch | 96 | 85% | *Das habe ich auch schon gesehen.....Kopftuch in fettem X7-BMW.* |
| Neger | 27 | 85% | *Was waren sie denn nun? Araber, Nafris, Neger?* |
| 😡 | 578 | 85% | *Findet diese Drecksau und entmannt sie an Ort und Stelle.* 😡😡😡 |
| 👊 | 132 | 83% | *im schutz seiner kettenhunde fühlt sich der stinker sicher!* 😡💪👊 |
| Muslime | 438 | 82% | *Muslime schlagen mit Stöcken auf Zuhörer ein* |
| Abschaum | 69 | 77% | *Einwanderung von Religiösem Abschaum = Riesenproblem* 💀💀💀💀 |
| Gefährder | 128 | 75% | *Gefährder und Kriminelle? Die sollte man überall hin abschieben.* |
| sofort | 344 | 69% | *Sofort müsste ihr ein Blitz in den Arsch fahren dieser Volksverräterin* |
| müssen | 1264 | 66% | *die Invasoren müssen lernen* |

**Table 2**. Sample keywords in German hate speech.

### 4.2.2 Word Co-occurrence

Not all of these words are necessarily hateful by themselves, e.g., *Flüchtlinge* is not a derogatory word, but can become so in combination with other words. To examine this, we used the Pattern toolkit to extract biased adjectives that co-occur with (precede) biased nouns, as in *kriminelle Flüchtlinge*. Thousands of biased adjective-noun pairs can be found almost exclusively in the hate speech data (~3,500), each of them usually occurring once or twice.[13]

Table 3 shows a sample of biased adjective-noun pairs, most of which are generalized and explicit (cf. section 3). For each pair, we have indicated the general targets occurring in most definitions of hate speech (i.e., race, gender, religion, ideology). Racist expressions typically portray dark-skinned people as dangerous and savage (e.g., predators, hordes), and/or ridicule Muslims by using negative stereotypes (e.g., bearded carpet kissers). Sexist expressions often decribe women as being naive and promiscuous (e.g., stupid sluts). Expressions that target religion depict Muslims as violent, misogynist, and unclean (e.g., stinking hate preachers). Expressions that target ideology depict left-wing voters as foolish and deceptive. Some expressions address a combination of targets, as in *weibliche Grüfri* ('Green fundamentalist chatterbox, an adaption of *Nafri*).

---

[13] Top 1,000 biased adjective-noun pairs: https://docs.google.com/spreadsheets/d/1gVfkxOzLiv47WH506eDseIji96vD2Q4ofFVTiVNUl4c

| ADJECTIVE | NOUN | RACE | GENDER | RELIGION | IDEOLOGY |
|---:|---|:---:|:---:|:---:|:---:|
| arbeitsscheue | Sozialschmarotzer | - | - | - | ✓ |
| bärtigen | Teppichknutscher | ✓ | - | ✓ | - |
| behaarte | Kanakenfotzen | ✓ | ✓ | - | - |
| blöde | Schlampe | - | ✓ | - | - |
| dreckige | Salafistenpack | - | - | ✓ | - |
| homosexuellen | Zellennachbarn | - | ✓ | - | - |
| linke | Heuchlerbande | - | - | - | ✓ |
| nordafrikanischen | Horden | ✓ | - | - | - |
| primitives | Negergesindel | ✓ | - | - | - |
| stinkenden | Hasspredigern | - | - | ✓ | - |
| schwarze | Trans-Muslima | ✓ | ✓ | ✓ | - |
| syrische | Bestien | ✓ | - | - | - |
| weibliche | Grüfri | - | ✓ | - | ✓ |
| widerliche | Systemkriecherin | - | - | - | ✓ |
| zwangsbekopftuchte | Mädchen | - | ✓ | ✓ | - |

**Table 3**. Sample word pairs in German hate speech.

Since these word pairs (but by extension also many compounds) are blatant in connecting a target to a derogatory word, they are particularly useful to expose two core mechanisms in hate speech: dehumanization and stereotyping.

**Dehumanization** can be defined as "processes associated with stripping groups or individuals of human 'essence' and processes that compare groups or individuals with nonhumans" (Goff, Eberhardt, Williams, & Jackson 2008:293). It has been used throughout history to pave the way for violence towards targeted out-groups (Haslam & Loughnan 2014:415). Some straightforward examples include comparisons to animals (e.g., apes, dogs, rats), infestations and diseases (e.g., parasites, pests), waste products (e.g., shit, trash), and implying a lack of intelligence or morality (barbarians, hordes).

**Stereotypes** are highly generalized beliefs about an out-group, often subconsciously rooted in intergroup relations with long histories, or consciously used to reinforce prejudices, for example to protect the in-group (Haslam & Loughnan 2014). Examples in the dataset include portrayals of politicians as being conniving and greedy, Muslims as screaming terrorists, immigrants as unemployed and loitering on street corners wearing jogging pants, and so on.

Finally, biased adjective-noun pairs are often combined with verbs that express aggression, such as *greifen* ('grab', ~1,000x), *schlagen* ('beat', 1,000x), *kämpfen* ('fight', 750x), *stechen* ('stab', 500x), and *klauen* ('rob' 250x), as in "Dunkelhäutiger Unterhosenmann geht plötzlich auf Renterin los, schlägt und tritt sie fast zu Tode" ('black man in underpants suddenly attacks elderly woman and nearly beats and kicks her to death').

### 4.2.3 Word Clusters

As noted earlier, users in the hate speech dataset predominantly seem to post tweets about immigration, crime, and politics. To assert this observation, we experimented with NLP techniques that automatically group words into meaningful categories. We used skip-grams (Mikolov, Sutskever, Chen, Corrado, & Dean 2013) and spherical *k*-means clustering (Hornik, Feinerer, Kober, & Buchta 2012) to compute three clusters ($k$=3) of the top 250 most biased words. A skip-gram represents the context of a word, i.e., those words that frequently precede or succeed it over all tweets. Using skip-grams, a clustering algorithm can then group words that more often occur in the same context into a single category (e.g., *gewalttätigen Muslimen / Terroristen / islamo-faschistoiden*).

Figure 1 shows a word cloud visualization of the three resulting clusters. The size of each word represents how often it appears in the hate speech dataset. For example, the second cluster groups words such as *Terroristen*, *Messer*, and *Islam*, which often appear in the same context in hate tweets, with *Islam* appearing most often (~1,400x), and *Messer* (~450x) and *Terroristen* (~200x) less frequently. We can roughly label the respective clusters as immigration, crime, and politics, although there is large semantic overlap between them. In effect, many utterances in the hate speech dataset attempt to defame refugees, implicate them in crimes, and blame politicians at the same time.

The accompanying table in Figure 1 shows ten representative words for each cluster, and the number of times they occur:

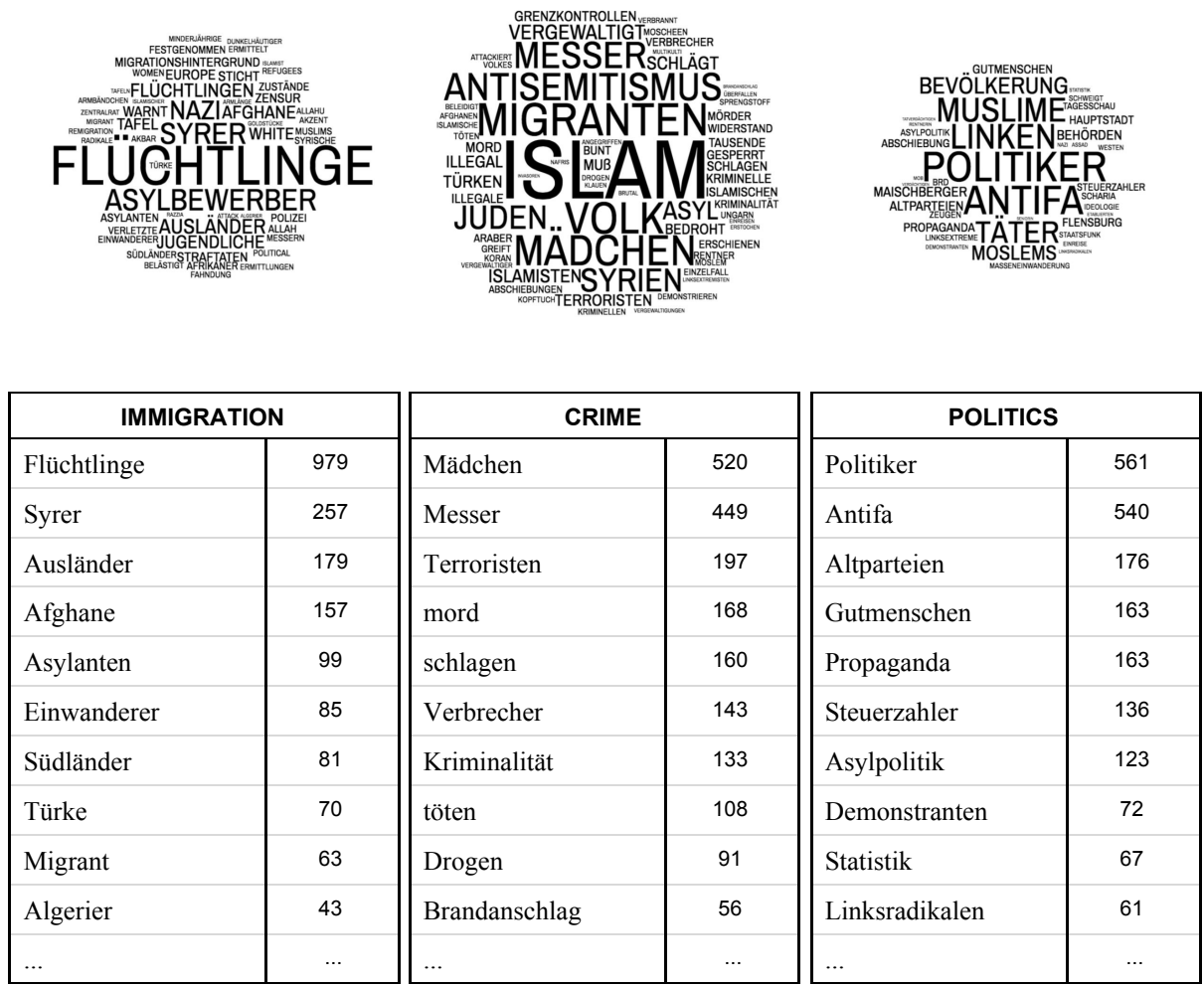

| IMMIGRATION | |
|---|---|
| Flüchtlinge | 979 |
| Syrer | 257 |
| Ausländer | 179 |
| Afghane | 157 |
| Asylanten | 99 |
| Einwanderer | 85 |
| Südländer | 81 |
| Türke | 70 |
| Migrant | 63 |
| Algerier | 43 |
| ... | ... |

| CRIME | |
|---|---|
| Mädchen | 520 |
| Messer | 449 |
| Terroristen | 197 |
| mord | 168 |
| schlagen | 160 |
| Verbrecher | 143 |
| Kriminalität | 133 |
| töten | 108 |
| Drogen | 91 |
| Brandanschlag | 56 |
| ... | ... |

| POLITICS | |
|---|---|
| Politiker | 561 |
| Antifa | 540 |
| Altparteien | 176 |
| Gutmenschen | 163 |
| Propaganda | 163 |
| Steuerzahler | 136 |
| Asylpolitik | 123 |
| Demonstranten | 72 |
| Statistik | 67 |
| Linksradikalen | 61 |
| ... | ... |

**Figure 1**. Sample word clusters in German hate speech.

### 4.2.5 Word Tree

The linguistic context of a word is telling of the world view of the author, and visualizing this context with a word tree can be a useful explanatory approach. To offer one example, Figure 2 shows the word tree for *Moslem*, where the size of each word represents how often it appears in the hate speech dataset. For example, *kriminell* and *radikal* appear more often in the vicinity before *Moslem* than *arme* ('poor'):

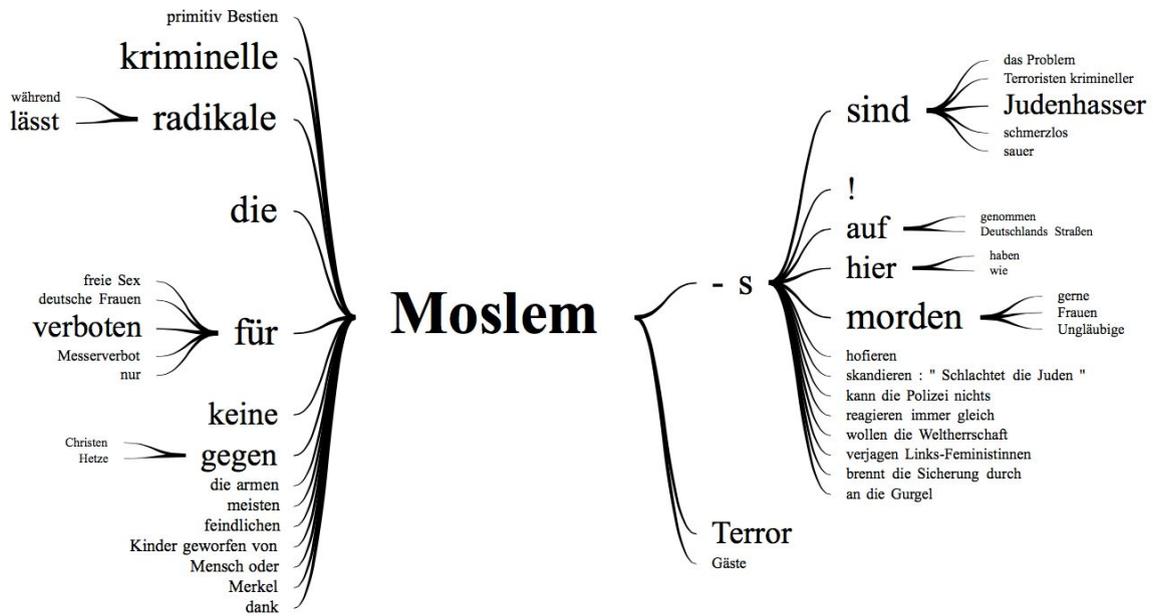

**Figure 2**. Sample word tree in German hate speech.

### 4.2.6 Word Polarity

Sentiment Analysis (Pang & Lee 2008) pertains to automatically detecting whether a text is factual or subjective, and if it is subjective, whether it is positive or negative (i.e., polarity). This task has a long history in NLP, and many systems have been developed for different languages, typically with a predictive performance that ranges between 75-85%. We used the SentiWS lexicon (Remus, Quasthoff, & Heyer 2010) for German. It assigns scores to words (e.g., *gut* = +0.37, *schlecht* = −0.77) that can be used to compute an average score for a given text. We computed the average score for all tweets resulting in about 32% of the hate tweets predicted as negative, against 22% of the safe tweets, or a 10% difference (Figure 3a).

We briefly examined the relation between polarity and demographic variables (region and gender) in hate tweets. The top 10 most often mentioned nationalities in hate tweets include German, African (mostly Malinese, Somali, Nigerian), Turkish, Israeli, Syrian, Russian, Afghan, Saudi, Austrian, closely

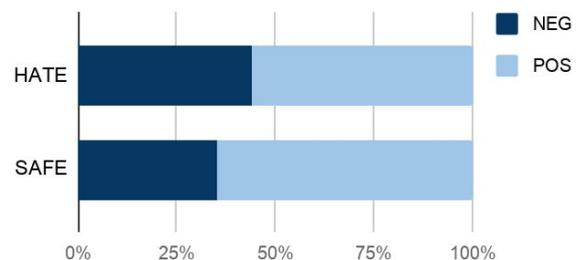

**Figure 3a**. Sentiment in German hate speech.

followed by Romanian, Iraqi, Moroccan (Figure 3b).[14]

Nearly all coverage of Somalis is negative (93%), as is coverage of Afghans (79%), Malinese (75%), Nigerians (73%), and also Turks (70%). The least negative content is about Austrians (48%) who are praised for their right-wing FPÖ victory in 2017.

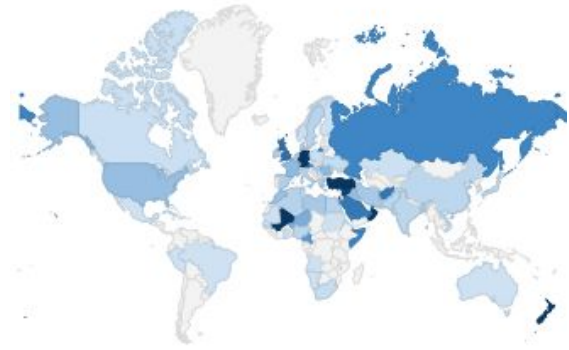

**Figure 3b**. References to region.

The hate tweets contain considerably more references (2:1 ratio) to women and girls (*Frauen*, *Mädchen*) than to men and boys (*Männer*, *Jungen*), and both more often in negative tweets. Note that such references may either target women or use them to target somebody else (Figure 3c).

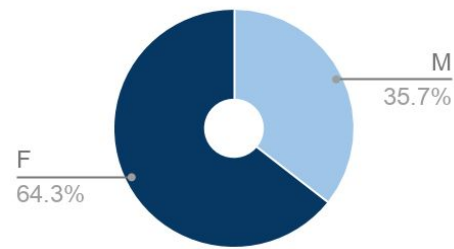

**Figure 3c**. References to gender.

## 4.3 Linguistic Creativity

Hate speech is not only marked by derogatory words or profanity. The dataset also shows that users can employ and bend language in a creative way to express protest against the current government, immigration policy, etc. This kind of creative language use typically involves lexis, but can also extend to other language levels such as syntax and grammar. We will restrict our brief observations to three typical cases of linguistic creativity in the dataset.

**Creative targeting**. The designations for immigrants and politicians are often inventive and display various linguistic techniques. These include creative compounds like *die Bundeskuh und ihr Idiotenstall* ('the Cowcellor and her stable of stupids'), *Politmaden-Bürgermeister* ('maggot mayor'), or *Maasdurchfallgesetz* ('Heiko Maas' NetzDiarrhea law'). Other cases involve blends such as *Erdowahn* ('Erdogoon') and *Religioten* ('religiots'). This is in line with findings for Italian data by Assimakopoulos, Baider, & Millar (2017:88).

---

[14] German appears most often because Germany is the central topic. Austrian is often mentioned because the Austrian immigration policy is seen as a role model. African nationalities, along with Muslims (i.e., Afghans, Syrians and Iraqis), are considered the main source of unwanted immigration. Turkish people are the largest group of immigrants in Germany and criticized for supporting Erdoğan's politics.

**Creative metaphors.** Metaphors are used within the argumentation to make an opinion more persuasive. One striking example targets Angela Merkel: "Ich denke das in Honeckers Versuchslabor nen Fenster angeklappt war. Und da kam die Merkel raus" ('I think a window in Honecker's lab was open and out came the Merkel'). Honecker and Merkel both originate from the eastern part of Germany. During the 1970s, with the Iron Curtain still in place, Erich Honecker was the leader of East Germany (Deutsche Demokratische Republik, DDR), after being imprisoned as a communist during the Nazi regime. Conspiracy theories, which are rife, suggest a connection between the two politicians. Metaphors are also used to target immigration policy, as in "Dämlicher Deutsche Politiker verlangen praktisch von Deutsche Bürger; während Dein Haus verbrennt, DU sollst Löschaktionen nicht bei Dir, sondern bei Nachbarhaus führen." This statement presents Germany as a burning house, while Germans put out fires in neighboring houses (probably a reference to other countries). In general, descriptions of politics and politicians are more creative than comments on the refugees' behavior. The latter tend to use widespread dehumanizing metaphors, particularly the parasite metaphor (cf. also Musolff 2015).

**Phraseological modification.** Another persuasion strategy consists of using phraseological units such as proverbs, sayings, or film titles, either in their canonical or a modified form. Proverbs lend themselves to the aim of persuasion, captivating "wisdom, truth, morals, and traditional views in a metaphorical, fixed and memorizable form and which is handed down from generation to generation" (Mieder 1993:24). The same goes for famous quotes. For example, "Der Krug geht so lange zum Brunnen bis er bricht" ('A jug is only useful in the well as long as it remains intact') encourages right-wing voters to keep fighting the system. Proverbs enable users to argue with easily understandable imagery that sums up complex challenges concisely. Here, they appear in various modified forms, as in "Verfassungsschutz beobachtet und wenn sie nicht gestorben sin, beobachten sie noch heute",[15] a modification of the classic fairy tale ending 'Und wenn sie nicht gestorben sind, dann leben sie noch heute' ('they lived happily ever after'). Another example, "Warum in die Ferne schweifen, wenn das Fremde liegt so nah?",[16] has the canonical form 'Warum in die Ferne schweifen, wenn das Gute liegt so nah?' ('all you need is right here').

---

[15] Free translation: 'The Office for the Protection of the Constitution is monitoring the case and will continue to do so forever'.

[16] Free translation: 'Why do we look at foreign countries if our country is full of foreigners?'

The advantage of phraseological modifications is that they have a highly associative potential due to the underlying original form, that they are attention-grabbing, allow for adaptations to specific contexts, present complex information concisely, and/or highlight the producer's intelligence (Jaki 2014:18).

## 4.4 Speech Acts

An analysis of tweets in terms of speech acts allows for describing the intention of what is uttered. Speech Act Theory (Searle 2008) distinguishes between different illocutions, i.e., meanings of utterances, and establishes assertives (e.g., 'The EU is working on hate speech regulation'), declarations ('I hereby fine you EUR 1500 for illegal hate speech'), directives ('Remove this profile!'), commissives ('We promise to remove the profile'), and expressives ('We wish we had less hate speech on our platform'). This variety of illocutions can also be found in hate speech (Sponholz 2018:65). There have been attempts for automatic speech act classification with Twitter data (e.g., Zhang, Gao, & Li 2011; Vosoughi & Roy 2016), but for English. Qualitative analysis of our data shows that hateful tweets are marked by some predominant (combinations of) speech acts and by indirect speech acts.

**Expressive speech acts** are the most aggressive and disparaging ones, as in "wenn ich diese Kasperle in Ihren roten Clownkostümen sehe kommt mir das kotzen" ('when I see these clowns in their red costumes, I could vomit'). They are usually accompanied by emojis, as in "mehr grüne 🤢🤢 sollten nicht sprechen dürfen! 💩😂💖" ('no more talk by greens'), where repeated emojis indicate distress, as in "DANKE FRAU MERKEL 👎👎👎". Expressive speech acts are used to vent negative emotions about politicians, political events, crime, etc. The majority is motivated by fear of change in Germany. Some are curses, like "Hoffentlich verrottet er in irgendeiner Ecke" ('hope he rots in some dark corner').

**Directive speech acts** are also common in the dataset, usually in combination with hashtags. For example, #AntiKap refers to the Antikapitalistisches Kollektiv, a network of right-wing militant protesters, and is used to remind members to gather for protests to disrupt the system, as in "Den nationalen Aufbau unterstützen! #NSjetzt #KapitalismusZerschlagen #AntiKap" ('Help us build our nation! National Socialism now, crush capitalism'). Another type of directive speech act calls to stop or reverse the intake of immigrants: "Obdachloser lebendig begraben! Man muss nicht lang überlegen, welche 'Kultur' hier wieder zugange war!

Abschaum finden und abschieben!" ('Homeless man buried alive! Doesn't take long to figure out which "culture" is at it again. Find the scum and deport it!').

**Assertive speech acts** are frequent in the dataset, but difficult to distinguish from expressive speech acts because retweets of news articles are often salted with an opinion. In cases like "Erzbischof beschreibt Migration als Waffe zur Islamisierung Europas" ('archbishop describes migration as a weapon for the Islamization of Europe'), the distinction is clear. But it is less clear in "Familiennachzug: De Maizière schlägt 'Vorab-Vereinbarung' vor und will 'Spaltung der Gesellschaft überwinden' Wenn ich schon das Transparent am Boden sehe, schwillt mir der Hals" ('Family reunification: De Maizière suggests 'prior agreement' and wants to 'overcome the division of society'. I get furious when I see the banner on the floor'), which is a combination of facts and opinions. An accompanying photo shows refugees with a banner stating that no human being is illegal.

**Commissive speech acts** are rare and usually constitute threats, e.g. "Erwischt es nur eine Freundin oder Bekannte von mir, leg ich den Pisser noch vorm Richter um 😡" ('if this happens to a friend of mine, I will kill the bastard in front of the judge').

**Indirect speech acts** occur particularly with the blurring of facts and opinions. The utterance "Almans werden auf öffentlichen Plätzen von Nafris abgezogen" ('blacks rob Germans at public places') is one example, being an expressive disguised as an assertive. It is not a real statement about the world, but part of a series of tweets by a user that exclusively spreads various racial stereotypes. Another example is "Ich wünsche gute Heimreise" ('safe travels home'), an amiable wish at first sight and as such an expressive speech act. In reality, the illocution is a demand, i.e., calling on refugees to go back to Syria. This becomes clear in the preceding part of the tweet, which claims that Islamic State in Syria has been annihilated.

# 5  Automatic Detection

## 5.1  In-domain Evaluation

Using Machine Learning (ML) techniques, we can take advantage of word bias and word co-occurrence to train a model that automatically detects hate speech. ML is a field of Artificial Intelligence aimed to develop algorithms that can 'learn by example'. When shown a 1,000 German texts and a 1,000 English texts, a machine learning algorithm can infer that *ü* and *ß* seldomly occur in English texts. Such cues can then be used to predict whether another text is written in German or in English. In the same way, we can use examples of hate speech and safe discourse to train a model to spot the differences. This works reasonably well on a coarse-grained level, although the task is much more challenging on a fine-grained level, i.e., predicting exactly what is illegal, exactly who is targeted, how to avoid false positives, and so on. To offer one example of a false positive: our model learned that *Diät* ('diet', as in weight-watching) is related to hate speech, since it was shown many hateful examples about *Diätenerhöhung* ('pay increase', where *Diät* means a politician's salary). This illustrates how deploying such automatic approaches can be problematic, falsely accusing weight watchers of proliferating hate speech is undesirable.

ML algorithms expect their input to be vectorized, i.e., given as a set of vectors, where each vector is a set of feature-weight pairs. In this case, each tweet could be a vector, the features could be words, and the weights could be word count. We used the Perceptron algorithm and character trigrams as features. A character trigram consists of three consecutive characters, e.g., *Flüchtling* = { `Flü, lüc, üch, cht, htl, tli, lin, ing` }. Character trigrams efficiently model linguistic variation such as spelling errors, word inflections, function words, etc. Modelling *Flüchtling* and *Flüchtlinge* as character trigrams hence ensures that they have several overlapping features – all except the *-nge* trigram.

Character trigrams (CH3) yield about 82% predictive performance.[17] We added additional features such as character bigrams (CH2), character unigrams (CH1 = punctuation marks, emojis), word unigrams (W1 = single words), and word bigrams (W2 = word pairs), boosting the performance by 2% (Table 4). We also removed usernames (@) to prevent overfitting. Overfitting means that the model memorizes training examples instead of discovering general patterns. Retaining usernames raises performance by 2%, but also means that the model becomes a blacklist of known usernames. It is possible that the model overfits on the writing style of the most prolific users, however.

---

[17] The weighted random baseline is 50%, i.e., a system randomly predicting hate or safe has a predictive performance of 50%.

| FEATURES | | | | | | ACCURACY | |
|---|---|---|---|---|---|---|---|
| CH3 | CH2 | CH1 | W1 | W2 | @ | P | R |
| ✓ | - | - | - | - | - | 83.12 | 82.19 |
| ✓ | ✓ | - | - | - | - | 83.02 | 82.61 |
| ✓ | ✓ | ✓ | - | - | - | 82.82 | 82.63 |
| ✓ | ✓ | ✓ | ✓ | - | - | 83.64 | 82.92 |
| ✓ | ✓ | ✓ | ✓ | ✓ | - | **84.21** | **83.97** |
| ✓ | ✓ | ✓ | ✓ | ✓ | ✓ | 86.23 | 86.09 |

**Table 4**. Overview of precision and recall for different features.

We trained the single-layer averaged Perceptron algorithm (Collins 2002) with an average F1-score of 84.21%.[18] The F1-score is the harmonic mean of recall and precision. Recall (R) is an estimate of how many hateful tweets the model is able to detect. Precision (P) is an estimate of how many tweets predicted as hateful are really hateful. For example, if the model marks every tweet that contains the word *Flüchtling* as hateful, its recall would be high but its precision would also be low, since many people talk about refugees without spreading hate speech. Recall and precision are obtained by training the model on a set of tweets, and then statistically testing its predictions on a different set. The results will include true positives, true negatives, false positives, and false negatives, where recall = `TP / (TP+FN)` and precision = `TP / (TP+FP)`. We used 10-fold cross validation with 50,000 hateful tweets and 50,000 safe tweets, meaning that 10 tests were performed with a different 9/10 of training data and 1/10 of testing data each, then averaging the results (see also Hartung, Klinger, Schmidtke, & Vogel 2017 for related prior work).

## 5.2 Cross-domain Evaluation

In ML, domain adaptation refers to the problem that a model trained on one kind of data will perform poorly on other kinds of data. For example, our model is trained on tweets and may perform poorly on blog posts because no training examples were ever provided for this kind of data. To estimate the scalability of our model, we tested its performance on a number of out-of-domain resources:

1. A hold-out set of the 1,000 most offensive examples in the hate speech dataset was composed by human annotators. This set was not used for training. We tested the trained model on the hold-out set, and about 92% of those tweets are correctly predicted as hateful.

2. A hold-out set of the 1,000 least offensive examples in the hate speech dataset was composed by human annotators. This set was not used for training. We tested the trained model on the hold-out set, and about 76% of those tweets are correctly predicted as safe.

3. We collected a 1,000 web pages from a far-right conspiracy website.[19] About 98% of these pages are correctly predicted as hateful.

4. We collected a 100 random articles from the German Wikipedia, which content is moderated for neutrality (NPOV).[20] About 90% of the articles are predicted as safe.

---

[18] Proof-of-concept in Python code: https://gist.github.com/tom-de-smedt/9c9d9b9168ba703e0c336ee0128ebae5

# 6 Discussion

In this section, we first summarize our insights from the descriptive parts of the paper, and the results of the ML experiment on automatic hate speech detection, before discussing the implications of our findings. Our main aim was to examine right-wing German hate speech on Twitter from a linguistic perspective. At the time of the 2017 German federal elections, we found a high number of tweets targeting Angela Merkel's immigration policy, and implicating refugees in criminal activities, often with the aim of spreading propaganda. The majority of these tweets use hateful language, especially to describe immigrants, but also politicians and other voters. As argued, a considerable part of the content is deceptive and dehumanizing, and some (but few) of the tweets are illegal according to German law. While hate speech is characterized by lexical elements like direct derogatory words or profanity, our analysis also shows that this does not apply to all hate tweets. Instead, a part of them are marked by creative language and more indirect offense, such as creative metaphors to designate targets, phraseological modifications, and indirect speech acts.

---

[19] http://wien.orf.at/news/stories/2901924

[20] https://de.wikipedia.org/wiki/Wikipedia:Neutraler_Standpunkt

Automatic detection systems can help to identify hate speech. We demonstrated that even a basic ML experiment can yield reliable results. Without doubt, the reliability of our model can be improved by more recent techniques (e.g., Deep Learning systems with embeddings). Another improvement would be taking into account the multimodality of tweets, such as text-image relations, since they are necessary to fully capture online political discourse. But finding ways to improve detection systems is only part of the challenge. The question is also where they can and should be applied. How much responsibility are we willing to delegate to AI for content moderation, especially for the removal of content? Even the best-performing systems to date for German hate speech are not entirely reliable,[21] especially since some types of hate speech (e.g., implicit) are difficult to detect. Deploying such systems without oversight risks producing a problematic amount of under- and overblocking. Hence, these systems should be used primarily to *support* human moderators of online forums.

We discussed how only a minority of tweets that may be perceived as hateful is actually illegal. Once reported, it is this fraction that will have to be removed according to NetzDG and the EU Code of Conduct. While other tweets may well be undesirable and morally questionable, they are protected by freedom of speech. So, if the majority of perceived hate speech is protected, do we actually need all the media coverage and the investments in AI detection tools? We argue that online hate speech is not just a matter of the law, but a societal challenge regarding common decency, human dignity, and democratic values. Consider that children in classrooms around the world are usually reprimanded when they say nasty things, illegal or not. This is because we have an interest in raising them to respect other people, in the hope that 'good behaviour' is reciprocated, fostering productive social interactions. As global voices in a digitalized society, we are still in our infancy in how we use social media for communication. Perhaps saying nasty things online should be reprimanded too, lest it becomes the new normal.

Hateful tweets are often not just expressions of opinion, but a tactic used by extremists attempting to normalize their ideology. This can inspire others to express themselves in a similar way (Langton 2018). In today's society, even political leaders broadcast expressions from the Nazi era, which only a few years ago would have provoked a scandal. When a member of German parliament speaks of *Machtergreifung* ('seizure of power'), promising to make things difficult for immigrants and dissidents,[22] this is now frowned upon but not considered exceptional anymore. Exactly how this escalation of online hate should be addressed remains an open question. In a society where all may speak, perhaps not by removing their voices. But history has also shown the consequences of letting everyone say whatever they want in any way they want.

---

[21] See GermEval Shared Task on the Identification of Offensive Language Online 2018 (Wiegand, Siegel, & Ruppenhofer 2018).

Some of these observations can also be linked to two essential principles in linguistics: (1) Language as an expression of how we think, of cognition, i.e., likely to indicate if a user is angry or not, and (2) language as also producing cognition. For example, if immigrants are repeatedly targeted with dehumanizing expressions, portrayed as infestations, this language use will spread. Also, the framing mechanism (see section 3) will influence the default values by which the cognitive frames of each individual are filled for concepts of immigration.[23] When characteristics like *kriminell* ('criminal'), *gewaltbereit* ('willing to use violence') or *arbeitsscheu* ('lazy') are repeatedly ascribed to refugees, the likelihood that we will think of refugees in terms of these characteristics increases, irrespective of reality. Similarly, if we habitually call politicians *Verräter* ('traitors'), *Kinderficker* ('pedos') or *Vollpfosten* ('idiots'), it will become more and more difficult to take them seriously.

To conclude, it is certainly true that we do not remove attitudes by removing the expression thereof. However, perhaps by being more restrictive of what morally questionable but legally acceptable content we accept online, we also have an influence on how extremist ideologies spread to the general population. In other words, accepting fewer offensive utterances online might help to reduce the reproduction of stereotypes and discrimination in the future.

---

[22] https://www.deutschlandfunk.de/die-afd-vor-den-wahlen-zwischen-umfragehoch-und-internem.724.de.html?dram:article_id=457285

[23] A frame is a data structure that is activated when we mention a word (for example), which has slots that must be filled with specific data; these slots are usually filled with some default values that can easily be adapted (cf. Minsky 1974:1f).